%% file: arxiv.tex
\documentclass[10pt,twocolumn,letterpaper]{article}

\usepackage[pagenumbers]{cvpr} %

\input{preamble}

\input{math_commands}

\usepackage{multirow}
\usepackage{algorithm}
\usepackage{algpseudocode}
\usepackage[subtle]{savetrees}

\definecolor{cvprblue}{rgb}{0.21,0.49,0.74}
\usepackage[pagebackref,breaklinks,colorlinks,allcolors=cvprblue]{hyperref}
\usepackage[accsupp]{axessibility}  %

\def\paperID{3479} %
\def\confName{CVPR}
\def\confYear{2025}

\title{ArchSym: Detecting 3D-Grounded Architectural Symmetries in the Wild}

\author{
    Hanyu Chen$^1$ \quad Ruojin Cai$^2$ \quad Steve Marschner$^1$ \quad Noah Snavely$^1$ \\[0.25em]
    $^1$Cornell University \quad $^2$ Kempner Institute, Harvard University
}

\begin{document}

\twocolumn[{%
\renewcommand\twocolumn[1][]{#1}%
\maketitle
\vspace{-2em}
\begin{center}
    \centering
    \includegraphics[width=\linewidth]{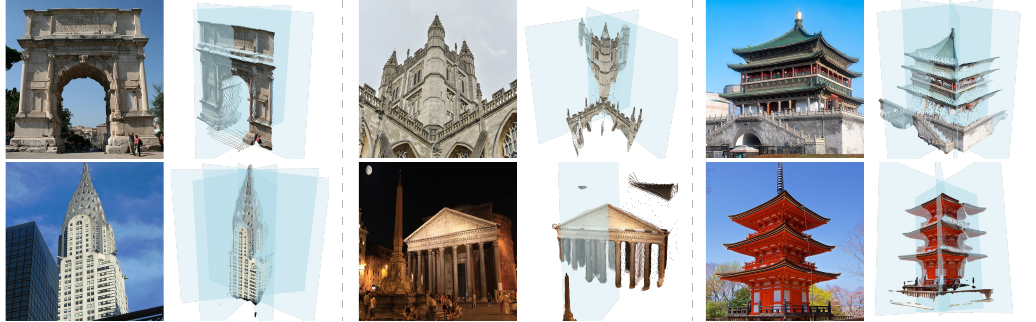}
    \vspace{-1.5em}
    \captionof{figure}{\textbf{Our method robustly detects 3D-grounded symmetries in challenging, in-the-wild images.} From a single RGB image (left in each pair), our model recovers dominant 3D symmetry planes (right) even when they are partially occluded or not directly visible. To train our model, we introduce a novel pipeline to automatically curate \archsym, a large-scale dataset of landmark symmetries. The results above are on images from landmarks unseen during training and highlight our model's strong generalization: it performs consistently on structures from different eras (first column); it is robust to extreme viewpoints and varying illumination (second column); and while trained primarily on Western landmarks, it generalizes effectively to diverse architectural styles (third column). \label{fig:teaser}}
\end{center}%
\vspace{1em}
}]

\begin{abstract}
Symmetry detection is a fundamental problem in computer vision, and symmetries serve as powerful priors for downstream tasks. 
However, existing learning-based methods for detecting 3D symmetries from single images have been almost exclusively trained and evaluated on object-centric or synthetic datasets, and thus fail to generalize to real-world scenes.
Furthermore, due to the inherent scale ambiguity of monocular inputs, which makes localizing the 3D plane an ill-posed problem, many existing works only predict the plane's orientation.
In this paper, we address these limitations by presenting the first framework for detecting \emph{3D-grounded reflectional symmetries} from single, in-the-wild RGB images, focusing on architectural landmarks. 
We introduce two key innovations: 
(1) a scalable data annotation pipeline to automatically curate a large-scale dataset of architectural symmetries, \archsym, from SfM reconstructions by leveraging cross-view image matching; 
and building on the dataset, (2) a single-view symmetry detector that accurately localizes symmetries in 3D by parameterizing them as signed distance maps defined relative to predicted scene geometry. We validate our symmetry annotation pipeline against geometry-based alternatives and demonstrate that our symmetry detector significantly outperforms state-of-the-art baselines on our new benchmark.
\end{abstract}

\section{Introduction}

\input{sections/introduction}

\section{Related work}

\input{sections/related_work}

\section{Dataset: \archsym}

\input{sections/dataset}

\section{Single-view symmetry detector}

\input{sections/method}

\section{Results}

\input{sections/results}

\section{Limitations}

\input{sections/limitations}

\section{Conclusion}

\input{sections/conclusion}

\vspace{1em}\noindent\textbf{Acknowledgments} This work was funded in part by the National Science Foundation (IIS-2212084) and by the Institute of Information \& Communications Technology Planning \& Evaluation (IITP) grant funded by the Korean Government (MSIT) (No. RS-2024-00457882, National AI Research Lab Project). We thank the authors of VGGT~\citep{wang2025vggt}, \ssb~\citep{li2025symmetry}, and \langevin~\citep{je2024robust} for assistance with model finetuning and running baselines.

{
    \small
    \bibliographystyle{ieeenat_fullname}
    \bibliography{main}
}

\clearpage
\appendix
\setcounter{figure}{0}
\setcounter{table}{0}
\setcounter{equation}{0}

\renewcommand{\thefigure}{A\arabic{figure}}
\renewcommand{\thetable}{A\arabic{table}}
\renewcommand{\theequation}{A\arabic{equation}}

\input{sections/suppl}

\end{document}

%% file: preamble.tex
\usepackage{microtype}

%% file: math_commands.tex
\usepackage{amsmath,amsfonts,bm}

\def\eqref#1{equation~\ref{#1}}

\def\1{\bm{1}}

\def\vn{{\bm{n}}}

\def\vp{{\bm{p}}}
\def\vq{{\bm{q}}}

\def\vt{{\bm{t}}}

\def\mK{{\bm{K}}}

\def\mR{{\bm{R}}}

\DeclareMathAlphabet{\mathsfit}{\encodingdefault}{\sfdefault}{m}{sl}
\SetMathAlphabet{\mathsfit}{bold}{\encodingdefault}{\sfdefault}{bx}{n}
\newcommand{\tens}[1]{\bm{\mathsfit{#1}}}

\def\tF{{\tens{F}}}
\def\tG{{\tens{G}}}

\def\gC{{\mathcal{C}}}

\def\gE{{\mathcal{E}}}

\def\gI{{\mathcal{I}}}

\def\gL{{\mathcal{L}}}

\def\gN{{\mathcal{N}}}

\def\gP{{\mathcal{P}}}

\def\gR{{\mathcal{R}}}
\def\gS{{\mathcal{S}}}

\def\sR{{\mathbb{R}}}

\DeclareMathOperator*{\argmin}{arg\,min}

\definecolor{blue}{HTML}{3F97CA}

\newcommand{\langevin}{\textsc{Langevin}\xspace}
\newcommand{\ssb}{\textsc{Reflect3D}\xspace}
\newcommand{\ssbshort}{\textsc{R3D}\xspace}
\newcommand{\ours}{\textsc{Ours}\xspace}
\newcommand{\ourscuration}{\textsc{Ours-Annotation}\xspace}
\newcommand{\direct}{\textsc{Direct}\xspace}
\newcommand{\directshort}{\textsc{Dir}\xspace}
\newcommand{\archsym}{ArchSym\xspace}

\newcommand{\wh}[1]{\widehat{#1}}

\newcommand{\planes}{\Pi}
\newcommand{\plane}{\pi}
\newcommand{\normal}{\vn}
\newcommand{\offset}{d}

\newcommand{\T}{\top}

\newcommand{\pixel}{x}
\newcommand{\pixelflipped}{x'}

\newcommand{\point}{\vp}
\newcommand{\points}{\gP}
\newcommand{\transformation}{T}

\newcommand{\image}{I}
\newcommand{\imageflipped}{I'}

\newcommand{\images}{\gI}
\newcommand{\imagesdef}{\{I_i\}}
\newcommand{\imagesflipped}{\gI'}
\newcommand{\imagesflippeddef}{\{I'_i\}}
\newcommand{\intrinsics}{\mK}

\newcommand{\extrinsics}{[\mR\mid \vt]}

\newcommand{\depthmap}{D}

\newcommand{\gtpoints}{\points}
\newcommand{\predpoints}{\wh{\points}}
\newcommand{\gtpoint}{\point}
\newcommand{\predpoint}{\wh{\point}}
\newcommand{\gtsdfmap}{\gS}
\newcommand{\sdf}{s}
\newcommand{\predsdfmap}{\wh{\gS}}
\newcommand{\gtsdf}{\sdf}
\newcommand{\predsdf}{\wh{s}}

\newcommand{\confmap}{\gC}
\newcommand{\conf}{c}

\newcommand{\featuremap}{\tF}
\newcommand{\query}{\vq}
\newcommand{\refinedquery}{\vq'}
\newcommand{\transformerdecoder}{\mathsf{Decoder}}
\newcommand{\adaln}{\mathsf{adaLN}}
\newcommand{\fusionblock}{\mathsf{FusionBlock}}

\newcommand{\fusedfeatureconditioned}{\tilde{\tG}}
\newcommand{\fusedfeature}{\tG}
\newcommand{\convhead}{\mathsf{ConvHead}}

\newcommand{\scale}{\bm\gamma}
\newcommand{\shift}{\bm\beta}

\newcommand{\npredsymmetries}{M}
\newcommand{\ngtsymmetries}{N}

\newcommand{\cost}{\gL}
\newcommand{\lone}{\ell^1}

\newcommand{\gtnormals}{\gN}
\newcommand{\gtnormalsvis}{\gN_\text{vis}}
\newcommand{\prednormals}{{\wh{\gN}}}
\newcommand{\prednormal}{{\wh{\normal}}}
\newcommand{\predoffset}{{\wh{\offset}}}
\newcommand{\predplane}{{\wh{\pi}}}
\newcommand{\reflection}{\gR}

\newcommand{\error}{\gE_{\text{dense}}}

\newcommand{\tp}{\text{tp}}
\newcommand{\fp}{\text{fp}}
\newcommand{\fn}{\text{fn}}
\newcommand{\ignore}{\text{nv}}
\newcommand{\matching}{\mathcal{M}}

%% file: sections/introduction.tex
Symmetry is a fundamental principle in nature and in human design, and it serves as a powerful prior for various computer vision tasks.
Symmetry is particularly useful for reasoning about occluded or partially observed geometry in 3D reconstruction and generation problems~\citep{wu2021rendering, wu2020unsupervised, li2025symmetry, yao2020front2back}. 
Symmetries also provide a canonical orientation for pose estimation~\citep{zhou2021nerd, shi2022symmetrygrasp} and help resolve ambiguities~\citep{zhao2023learning, merrill2022symmetry}.
Detecting symmetries in visual data is a long-standing problem with foundational work dating back several decades~\citep{wolter1985optimal, atallah1985symmetry}. 
Early approaches relied on geometric heuristics and handcrafted features to identify symmetric patterns in images or 3D models.
While effective in controlled settings, these classical methods struggle with real-world scenarios.
Recent learning-based methods~\citep{shi2020symmetrynet, zhou2021nerd, li2025symmetry} have achieved impressive results that generalize better than their handcrafted predecessors.

Despite this progress, existing methods for single-view 3D symmetry detection are almost exclusively trained and evaluated on object-centric datasets, such as ShapeNet~\citep{chang2015shapenet} and Objaverse~\citep{deitke2023objaverse}, which contain clean, pre-segmented objects without complex backgrounds. 
Consequently, their performance degrades significantly when applied to in-the-wild scenes featuring complex environments, varying illumination, and occlusions, leaving the problem of 3D symmetry detection in real-world environments largely unsolved.

In this work, we present a novel pipeline for predicting scene geometry and 3D reflectional symmetries grounded in the predicted geometry
from single images of \emph{in-the-wild} scenes. We focus specifically on architectural scenes, as their man-made designs often contain symmetrical structures. 
We first introduce a scalable method for curating a dataset of architectural images labeled with 3D reflectional symmetries. 
Our method is inspired by the ``doppelganger'' problem in 3D reconstruction \citep{cai2023doppelgangers, xiangli2025doppelgangers++}, where image matchers fail to distinguish physically distinct, but visually similar structures. 
We leverage this property of feature matchers to automatically annotate symmetries from structure-from-motion (SfM) reconstructions. 
Building on this data, we introduce a novel single-view 3D symmetry detector that parameterizes reflectional symmetries as \emph{signed distance maps} defined relative to the predicted scene geometry.
This parameterization resolves the scale ambiguity inherent to single-view symmetry detection methods, enabling accurate 3D localization of symmetry planes.

The main contributions of our work are as follows:
\begin{itemize}
\item We introduce a new scalable pipeline for automatically curating 3D symmetry annotations, yielding \archsym, a large-scale dataset of in-the-wild landmark symmetries.
\item We present the first end-to-end model that detects 3D-grounded symmetries from real-world images by parameterizing them relative to the predicted scene geometry.
\item We establish a new benchmark for this task and demonstrate that our model significantly outperforms state-of-the-art methods, even when they are finetuned on \archsym.
\end{itemize}

%% file: sections/related_work.tex
\noindent\textbf{Symmetry detection.} Prior work in symmetry detection can be categorized along two primary axes: the input modality (e.g., RGB images, RGB-D scans, or 3D models) and the output space (2D vs.\ 3D). 
Here, we review key methods that are most relevant to our work. For comprehensive surveys, we refer the reader to \citet{mitra2013symmetry} for classical methods and \citet{funk20172017} for learning-based methods.

Many early methods for symmetry detection from images focus on identifying symmetries in front-facing objects, where 3D reflectional planes reduce to 2D lines of symmetry in the image plane~\citep{kiryati1998detecting, loy2006detecting, tsogkas2012learning, nagar2017symmmap, elawady2017wavelet, cicconet2017finding}. Subsequent work extends this idea to dense heatmaps that highlight centers of rotation and lines of reflection for unconstrained, real-world images~\citep{funk2017beyond}. 
While effective for 2D tasks, these approaches inherently lack the ability to localize symmetries in 3D.

More recently, machine learning has led to significant progress in 3D symmetry detection. 
Such approaches can be broadly classified by input modality: RGB images~\citep{zhou2021nerd, lin2021nerd++, li2025symmetry}, RGB-D scans~\citep{shi2022symmetrygrasp, shi2020symmetrynet, shi2022learning}, or 3D models and point clouds~\citep{li2023e3sym, gao2020prs, je2024robust, zhang2023single}. 
Most related to our work are NeRD/NeRD++~\citep{zhou2021nerd, lin2021nerd++} and Reflect3D~\citep{li2025symmetry}, which detect reflectional symmetries from single-view RGB images. 
The former proposes an iterative refinement strategy that reflects image features in 3D to identify symmetry planes; the latter leverages foundation models to regress plane parameters. 
However, a key limitation of these methods is that they are trained almost exclusively on object-centric data, and their performance degrades significantly in complex, in-the-wild scenes.

\medskip\noindent\textbf{Symmetry for 3D reconstruction.} Symmetry is a powerful prior for reconstructing 3D geometry, particularly from limited data.
Early work like that of \citet{sawada2014detecting} derives geometric conditions under which 2D object contours imply a 3D mirror symmetry to constrain the ill-posed problem of monocular shape recovery. 
\citet{koser2011dense} propose detecting a 3D symmetry plane and performing dense reconstruction by matching features between an image and its reflected counterpart.
In particular, this idea suggests that image matching can be a useful tool for symmetry extraction.

More recently, symmetry has also been leveraged as a cue for learning-based 3D reconstruction and generation models.
\citet{wu2020unsupervised}, for example, use a symmetry prior to disentangle depth, albedo, and viewpoint in single-view 3D reconstruction. 
\citet{li2025symmetry} detect and aggregate symmetry planes from multiple views to guide mesh generation, ensuring global consistency from partial observations.

\medskip\noindent\textbf{Symmetry ambiguities in 3D reconstruction.} Conversely, symmetry and other forms of structural repetition are often problematic in structure-from-motion (SfM)~\citep{schonberger2016structure}. 
The visual similarity between distinct but symmetric parts of a scene leads image matchers to produce incorrect correspondences. These illusory matches, or ``doppelgangers''~\citep{cai2023doppelgangers, xiangli2025doppelgangers++}, can introduce significant errors in 3D reconstruction, particularly for large-scale architectural scenes where such ambiguities are common. 
However, this phenomenon also suggests that these incorrect, yet \emph{geometrically consistent}, matches could be a valuable signal for discovering landmark symmetries.

\medskip\noindent\textbf{Automated symmetry annotation.} Learning-based methods have proven to excel in symmetry detection, but curating a dataset of ground truth 3D symmetry annotations, especially for real-world scenes, is challenging. 
Prior methods that use synthetic and object-centric data have relied on pre-aligned 3D models~\citep{zhou2021nerd}, iterative closest point algorithms~\citep{li2025symmetry}, and manual labeling~\citep{shi2020symmetrynet, shi2022learning}. 
Several unsupervised~\citep{li2023e3sym, gao2020prs} and training-free~\citep{je2024robust} methods have also been proposed to automatically generate symmetry annotations from point clouds. 
However, these geometry-based methods often struggle with the noisy and incomplete point clouds produced by SfM on in-the-wild scenes. 
By operating solely on geometry, they also discard visual information in the original images. 
This motivates the need for alternative data annotation methods that can 
robustly identify symmetries in real-world reconstructions.

%% file: sections/dataset.tex
\label{sec:dataset_curation}

The goal of our work is to predict the plane parameters of \emph{global reflectional symmetries} from a single input image. 
Formally, given an image \(I\), we aim to learn a mapping from the image to a set of symmetry planes, \(\planes = \{\plane_k\}_{k=1}^{\ngtsymmetries}\), represented in the camera coordinate system. Each plane is defined by a unit normal \(\normal \in S^2\) and its signed distance to the origin, or offset, \(\offset \in \mathbb{R}\).
We define global reflectional symmetries as planes of reflection that apply to an entire landmark structure or to a complete facade. As photographs often capture single, symmetric facades of otherwise asymmetric buildings, treating these as global symmetries reflects this common real-world scenario and provides a well-defined target for our model.

A large-scale dataset of real-world 3D symmetries is essential for training a symmetry detector for in-the-wild images.
However, defining real-world symmetries is inherently subjective, as landmarks inevitably contain minor imperfections that break perfect symmetry. To resolve this, we ground our definition in \emph{human perception}: a structure is considered symmetric if its distinct surfaces are visually indistinguishable, a phenomenon known as \emph{perceptual aliasing}. This ensures we capture dominant structural symmetries while naturally ignoring local architectural variations.
Guided by this perceptual definition, we introduce an automated pipeline that extracts symmetry annotations directly from structure-from-motion (SfM) reconstructions of landmark scenes.

\begin{figure}[t]
    \centering
    \includegraphics[width=\linewidth]{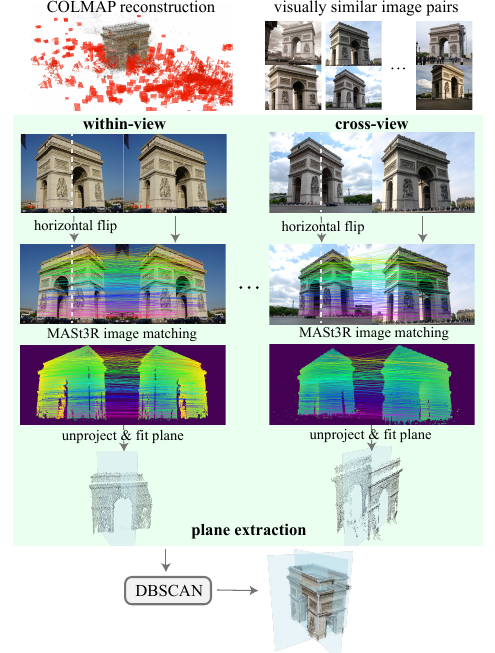}
    \vspace{-2em}
    \caption{\textbf{Overview of our automated pipeline for extracting symmetry annotations.} 
    We visualize \emph{within-view} (left) and \emph{cross-view} (right) matching on image pairs sampled from an SfM reconstruction. 
    For each pair, we horizontally flip one image, find dense matches with the other image via MASt3R~\citep{duisterhof2024mast3r}, unproject matched pixels to 3D points using depth maps, and fit a plane to the resulting point pairs. 
    The final symmetry planes annotations are then determined by clustering candidate planes with DBSCAN~\citep{ester1996density}.}
    \label{fig:matching}
    \vspace{-1em}
\end{figure}

Our pipeline is motivated by two observations. The first is a classical technique in 3D reconstruction: matching an image to its own reflected counterpart is an effective way to extract symmetries visible from the image \citep{koser2011dense} (\emph{within-view} matching). The second is an insight from the ``doppelganger'' problem: image matchers often find incorrect, yet geometrically consistent, matches between visually similar but physically distinct structures \citep{cai2023doppelgangers, xiangli2025doppelgangers++}. 
We find that within-view matching tends to only discover dominant symmetries, such as a reflection across a main facade. Therefore, we leverage the insight from the doppelganger problem and additionally match each image against the reflected versions of \emph{other}, visually similar images from the scene (\emph{cross-view} matching).
This allows us to recover symmetries that are not apparent or fully visible from any single viewpoint. For example, as visualized in Figure~\ref{fig:matching}, within-view matching on the Arc de Triomphe extracts the reflection across its main facade, while cross-view matching recovers the front-to-back reflection, which is less obvious from a single image.

\begin{figure*}
    \centering
    \includegraphics[width=\linewidth]{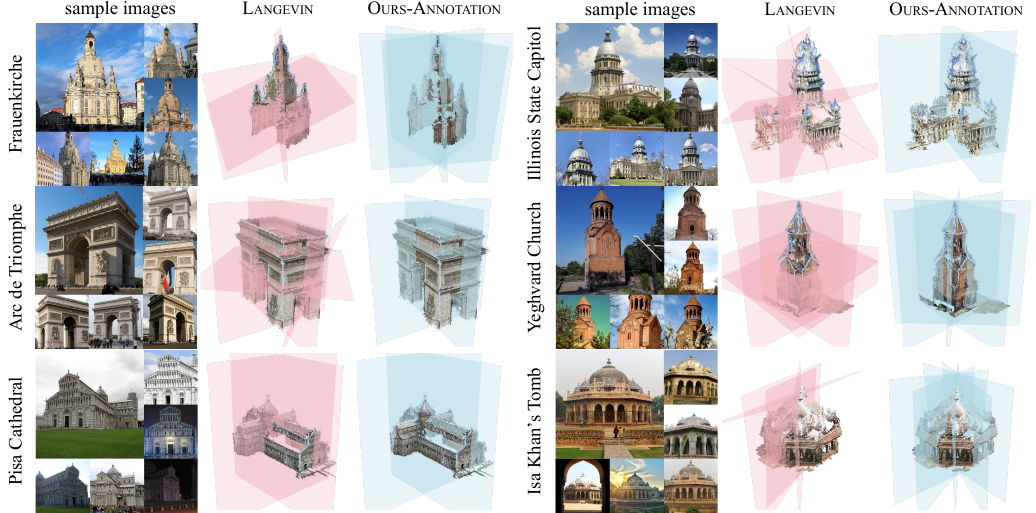}
    \vspace{-1.75em}
    \caption{\textbf{Visualization of automated symmetry plane annotations.} We run \langevin~\citep{je2024robust} and our symmetry extraction pipeline on six scenes from MegaScenes. Extracted planes are visualized with a dense point cloud from COLMAP~\citep{schonberger2016structure, schoenberger2016mvs}. For \langevin, the dense point cloud from COLMAP is used as input. For \ourscuration, sampled pairs of input images and depth maps are used as input. We observe that \langevin, as a purely geometry-based method, is highly sensitive to incomplete point clouds (e.g. Frauenkirche, Isa Khan's Tomb) and detects architecturally implausible planes (e.g., horizontal plane on the Arc de Triomphe, misaligned plane on the Pisa Cathedral). In contrast, \ourscuration extracts semantically-correct symmetries. All planes are visualized without manual filtering or postprocessing.
    }
    \label{fig:curation_comparison}
    \vspace{-1em}
\end{figure*}

\subsection{Data and preprocessing}
To create \archsym, we select 93 landmark scenes from MegaScenes, a large dataset of in-the-wild landmark image collections~\citep{tung2024megascenes}. 
For each scene, given its set of images \(\images = \imagesdef\), we produce a structure-from-motion (SfM) reconstruction of the scene by running MASt3R-SfM~\citep{duisterhof2024mast3r}, along with  Doppelganger++~\citep{xiangli2025doppelgangers++} to correct reconstruction errors. 
The reconstruction yields an intrinsic camera matrix, extrinsic pose, and estimated depth map for each image. 
We also produce a set of horizontally reflected images \(\imagesflipped=\imagesflippeddef\). 

\subsection{Symmetry plane annotation}
As shown in Figure~\ref{fig:matching}, the process for generating a single symmetry annotation from a pair of images is as follows:
\begin{enumerate}
    \item \textbf{Image pair selection.} For each image \(\image_i\), in addition to its own reflection \(\imageflipped_i\), we sample a subset of visually similar reflected images \(\mathcal J_i \subset \imagesflipped\) 
    based on ASMK similarity~\citep{tolias2013aggregate}.

    \item \textbf{Image matching.} Pairing image \(\image_i\) with flipped image \(\imageflipped_j \in \mathcal J_i \cup\{\imageflipped_i\}\), we run an image matcher, MASt3R~\citep{leroy2024grounding}, to find 2D correspondences \(\mathcal{M} = \{(\pixel^k_i, \pixelflipped^k_j)\}\).

    \item \textbf{3D point unprojection.} 
    For each 2D correspondence \((\pixel^k_i, \pixelflipped^k_j)\), pixel \(\pixel^k_i\) is unprojected to a 3D point \(\point^k_i\) using its depth and camera parameters from the reconstruction.
    The corresponding 3D point \(\point^k_j\) is similarly computed by unprojecting the original, un-flipped pixel coordinate of \(\pixelflipped^k_j\). This yields two sets of corresponding 3D points, \(\points_i\) and \(\points_j\), approximately related by a reflectional symmetry.

    \item \textbf{Symmetry plane fitting.} We estimate the parameters of a candidate symmetry plane \(\plane^* = (\normal^*, \offset^*)\) from the 3D point correspondences
    by minimizing the sum of squared distances between points in \(\points_i\) and reflections of corresponding points in \(\points_j\):
    \begin{equation}
        (\normal^*, \offset^*) = \argmin_{\|\normal\|=1, \offset} \sum_{k} \|\point^k_i - \reflection_{\normal,\offset}(\point^k_j) \|^2.
    \end{equation}
    where \(\reflection_{\normal,\offset}\) is the reflection operator for a symmetry plane with normal \(\normal\) and offset \(\offset\).
\end{enumerate}

\subsection{Plane aggregation and verification}
Each image pair yields a potentially noisy candidate plane. We aggregate thousands of planes from each scene and use DBSCAN~\citep{ester1996density} to cluster these candidate planes.
The centers of large clusters are kept as high-confidence symmetry annotations.
For the final dataset, we manually inspect these high-confidence planes
and filter out any incorrect or local symmetries and add global symmetries that our automated process may have missed. The resulting planes are treated as ground truths for training our symmetry detector.

\subsection{Symmetry annotation results}

\noindent\textbf{Dataset statistics.} Our final \archsym dataset consists of 93 landmark scenes and a total of 34,177 images. The distributions of images and annotated symmetries per scene are shown in Figure~\ref{fig:dataset_stats}. 
Most scenes contain one or two symmetries, while a few contain four (e.g., clock towers) or up to eight (e.g., octagonal buildings) symmetries. 
The images contribute a wide range of challenges, including partial views, occlusions, and varying camera parameters and illumination. 
The dataset provides a rigorous benchmark for evaluating symmetry detection across a wide range of real-world image conditions. 

\medskip\noindent\textbf{Qualitative comparisons.} 
To validate the effectiveness of the symmetry extraction method used to build \archsym, we compare the quality of our extracted symmetry annotations to those produced by a recent state-of-the-art geometry-based method that uses Riemannian Langevin dynamics to detect reflectional symmetries \citep{je2024robust}, or \langevin  for short. 
As ground truth is subjective for this task, the comparison is primarily qualitative (Figure~\ref{fig:curation_comparison}). 
We provide dense SfM point clouds as input to \langevin. 
All visualized planes are direct outputs of the two methods without manually filtering.

For each scene, we visualize sample images and the dense point cloud from COLMAP overlaid with the symmetry planes detected by \langevin and our method. We note several failure modes of the geometry-based approach. 
\langevin is highly sensitive to incomplete reconstructions and fails to identify symmetries where points to one side of the symmetry plane are largely missing (e.g., Isa Khan's Tomb, Frauenkirche). 
It often prioritizes the coarse shape of the point cloud over the underlying architectural semantics. For example, on the Pisa Cathedral, \langevin incorrectly places a symmetry plane halfway along the building's length, and on cuboid-like silhouettes (e.g., Arc de Triomphe), it detects horizontal planes that are architecturally implausible.

In contrast, our method consistently extracts semantically correct two-way and four-way symmetries for landmarks across a wide range of architectural styles. 
For more challenging landmarks such as Isa Khan's Tomb, which has an eight-way symmetry, our method successfully extracts five of these reflectional symmetries via image matching, and the remaining three symmetries can be robustly estimated from the shared intersection line of the extracted symmetries during postprocessing. 
Moreover, for scenes like the Pisa Cathedral, our method correctly identifies the multiple reflectional symmetries at the center of the distinct facades.

%% file: sections/method.tex
\begin{figure}[t]
    \centering
    \includegraphics[width=\linewidth]{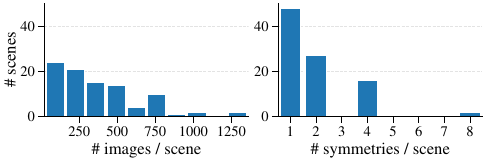}
    \vspace{-1.75em}
    \caption{\textbf{Statistics of the \archsym dataset,} showing the distribution of the number of images available (left) and the number of global symmetries annotated (right) in each scene.}
    \label{fig:dataset_stats}
    \vspace{-1em}
\end{figure}

\label{sec:symmetry_detector}
Using our newly curated \archsym dataset, we train a model to predict 3D symmetry planes from a single RGB image. We choose to finetune a 3D foundation model, VGGT~\citep{wang2025vggt}, adding a symmetry plane prediction head to its backbone, which is pretrained on diverse 3D geometry tasks. We elaborate on the model architecture in Section~\ref{sec:prediction_head}.

A key challenge in single-view 3D symmetry detection is \emph{scale ambiguity}, which makes grounding symmetry planes in 3D an ill-posed task.
Existing methods~\citep{zhou2021nerd,li2025symmetry,shi2020symmetrynet,shi2022learning} address scale ambiguity only partially---either by only predicting plane normals and relying on downstream optimization to localize them in 3D, or by using RGB-D inputs that provide a reference scale.
Our choice of finetuning VGGT, which already predicts scene geometry, circumvents this issue: the predicted point map provides a natural, scale-consistent coordinate frame for grounding 3D symmetry. 

However, a straightforward two-stage baseline of running a geometry-based symmetry detector on this point map output often fails in practice, since VGGT predicts geometry for only the \emph{visible} parts of the scene, resulting in incomplete point clouds that often include irrelevant geometry (e.g., other buildings).
Therefore, we instead allow our new symmetry head to implicitly reason about symmetries by operating directly on the frozen VGGT backbone features. 
As we describe next, we parameterize reflectional symmetries in relation to this predicted geometry, ensuring that our predictions are both 3D-grounded and consistent with the underlying scene geometry.

\begin{figure*}
    \centering
    \includegraphics[width=\linewidth]{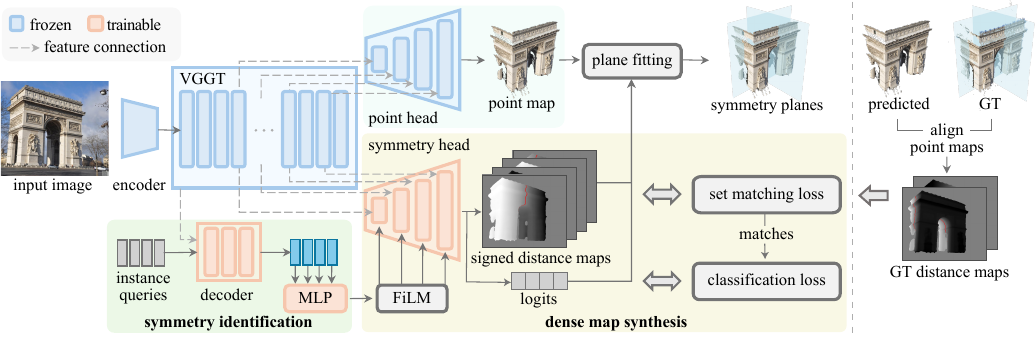}
    \vspace{-1.75em}
    \caption{\textbf{Overview of our single-view symmetry detector architecture.} A frozen VGGT~\citep{wang2025vggt} backbone first extracts features from a single input image. The features are processed by a transformer decoder with learnable instance queries to identify symmetries. A lightweight MLP generates conditioning parameters from the resulting instance features. Then, a DPT-style~\citep{ranftl2021vision} prediction head fuses multi-layer features to generate signed distance maps, with conditioning parameters injected via FiLM layers~\citep{perez2018film}. The model also predicts classification logits for extracting valid symmetries at inference time. The final 3D-grounded symmetry planes are recovered by solving an optimization problem over the predicted signed distance maps and the point map from VGGT's frozen point head. In the visualized signed distance maps, lighter pixels indicate positive values and darker pixels indicate negative values. Pixels with values close to zero are highlighted in red.}
    \label{fig:prediction_head}
    \vspace{-1em}
\end{figure*}

\subsection{Symmetry prediction as a signed distance map}

To provide a scale-consistent supervision signal, rather than directly regressing plane parameters, we formulate the problem as a dense prediction task: for each pixel, the model predicts the signed distance from that pixel's corresponding \emph{predicted} 3D point to the ground truth symmetry plane.

Given a predicted point map \(\predpoints = \{\predpoint^k\}\) from the frozen point head and a ground truth point map \(\gtpoints = \{\gtpoint^k\}\), we first compute a similarity transformation \(\transformation : \sR^3 \to \sR^3\) that aligns \(\gtpoints\) to \(\predpoints\). 
We also apply 
\(\transformation\) to the ground truth symmetry plane to align it to the predicted point map. 
Using the aligned plane \(\plane = (\normal, \offset)\), we compute a signed distance map \(\gtsdfmap = \{\gtsdf^k\}\) as
\begin{equation}
    \gtsdf^k = \normal^\T \predpoint^k + \offset.
\end{equation}

This signed distance map is treated as the \emph{ground truth} for training purposes\footnote{Crucially, this signed distance map is derived from the network's own \emph{predicted} geometry as opposed to the ground truth geometry, which allows the model to make \emph{self-consistent} predictions.}. We note that these signed distance maps remain constant throughout training, since they correspond to the fixed 3D pointmaps predicted by the fixed VGGT head. 
This provides a stable, scale-consistent supervision signal that naturally resolves inherent scale ambiguities.
We visualize examples of signed distance maps in Figure~\ref{fig:prediction_head}.

\subsection{Multiple symmetry plane prediction head}
\label{sec:prediction_head}
Detecting multiple symmetries within a single image requires both global reasoning to identify potential symmetry planes and pixel-level precision to accurately localize each plane in 3D. 
Accordingly, our symmetry plane prediction head is designed as a two-stage architecture.
In the first stage, a transformer-based~\citep{vaswani2017attention} module identifies a set of potential symmetry planes, while in the second stage, a dense prediction head regresses the actual signed distance maps. An overview of the architecture is shown in Figure~\ref{fig:prediction_head}.

\medskip\noindent\textbf{Symmetry identification module.} The first stage of the network performs high-level reasoning on the \emph{final-layer} feature map \(\featuremap^\text{final}\) from the frozen VGGT backbone. A set of \(\npredsymmetries\) learnable instance queries \((\query_1, \dots, \query_\npredsymmetries)\) are refined by a lightweight transformer decoder that allows them to attend to frozen backbone features:
\begin{equation}
    (\refinedquery_1, \dots, \refinedquery_\npredsymmetries) = \transformerdecoder((\query_1, \dots, \query_\npredsymmetries), \featuremap^\text{final}).
\end{equation}
Intuitively, each refined instance feature vector encodes information about a potential symmetry plane in the scene.

\begin{figure*}
    \centering
    \includegraphics[width=\textwidth]{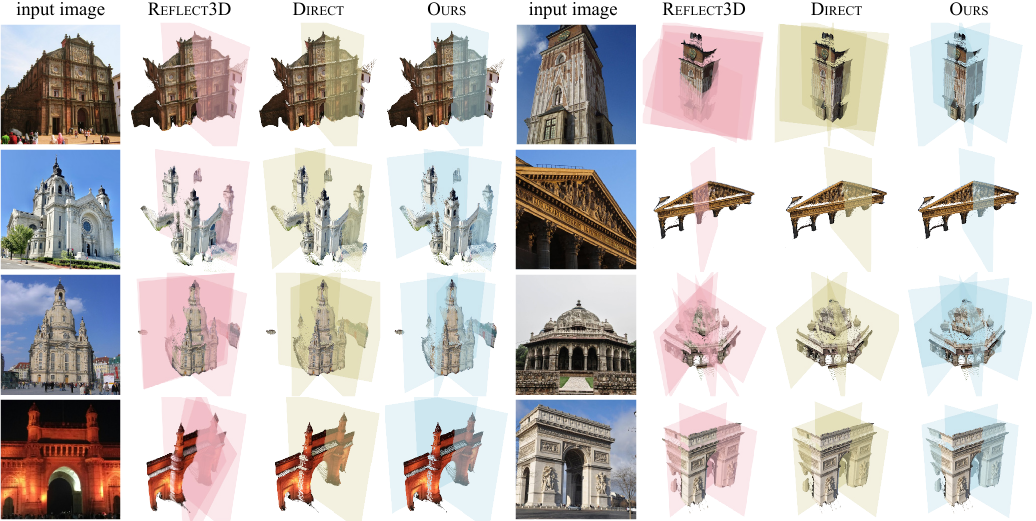}
    \vspace{-1.75em}
    \caption{\textbf{Qualitative comparison of single-view symmetry detection results.} Input images are sampled from eight different test scenes. Since \ssb~\citep{li2025symmetry} does not predict plane offsets, for visualization purposes, we use the point closest to the center of the landmark on the corresponding ground truth plane as an anchor point for \ssb's predicted normals. Point clouds shown are predicted by VGGT~\citep{wang2025vggt}. We observe that \ssb often misses partially visible symmetries and produces redundant detections, \direct predicts planes with more accurate normals but are often misaligned with the scene geometry, while \ours consistently predicts planes with accurate orientation and alignment. We encourage zooming into the figure to see differences in plane orientation and alignment in detail.}
    \label{fig:detection_qualitative}
    \vspace{-0.75em}
\end{figure*}

\medskip\noindent\textbf{Dense map synthesis module.} The second stage synthesizes the final dense maps using a dense prediction transformer (DPT) head~\citep{ranftl2021vision}.
We propose a lightweight extension to the DPT head that allows it to predict multiple instances of symmetry planes: Before each of the four feature fusion blocks, we inject instance-specific information using a feature-wise linear modulation (FiLM) layer~\citep{perez2018film}.

Scale and shift \((\scale^l_i, \shift^l_i)\) vectors used for the FiLM layers are regressed from refined instance features \(\refinedquery_i\) from the previous stage through a small MLP.
We follow \citet{peebles2023scalable} and apply adaptive layer normalization (adaLN) before each conditioning layer. Each fusion stage takes the current fused feature \(\fusedfeature^l_i\) and the feature map \(\featuremap^l\) from the VGGT backbone and combines them via a RefineNet-based feature fusion block~\citep{ranftl2021vision, lin2017refinenet}:
\begin{align}
    \fusedfeatureconditioned^l_i &= (1 + \scale^l_i) \cdot \adaln(\fusedfeature^l_i) + \shift^l_i, \\
    \fusedfeature^{l+1}_i &= \fusionblock(\fusedfeatureconditioned^l_i, \featuremap^l).
\end{align}
The output of the final fusion block is passed through a lightweight convolutional head to produce a set of predicted signed distance maps and confidence maps:
\begin{equation}
    (\predsdfmap_i, \confmap_i) = \convhead(\fusedfeature^\text{final}_i).
\end{equation}

\medskip\noindent\textbf{Bipartite matching loss.} Given ground truth distance maps \(\{\gtsdfmap_1, \dots, \gtsdfmap_\ngtsymmetries\}\), predicted distance maps \(\{\predsdfmap_1, \dots, \predsdfmap_\npredsymmetries\}\), and confidence maps \(\{\confmap_1, \dots, \confmap_\npredsymmetries\}\), we compute pairwise matching costs between all ground truths and predictions and find an optimal matching using the Hungarian matching algorithm~\citep{kuhn1955hungarian}, following prior work (e.g. DETR~\citep{carion2020end}). Similar to the confidence-aware point prediction loss proposed by \citet{wang2024dust3r}, the pairwise matching cost is computed as
\begin{equation}
    \cost_{ij} = \sum_{k} \conf^k_i \cdot \lone(\predsdf^k_i, \gtsdf^k_j) - \alpha \log\conf^k_i,
\end{equation}
where \(\alpha\) is a hyperparameter that controls the regularization strength. 
Given an optimal matching \(\matching\), the final loss is computed as the mean cost over matched pairs:
\begin{equation}
    \cost = \frac{1}{|\matching|} \sum_{(i, j) \in \matching} \cost_{ij}.
\end{equation}

\medskip\noindent\textbf{Classification loss.} To filter predicted symmetry planes, we pass the final feature maps through a small MLP to produce classification logits. Predicted planes matched to ground truth planes receive a positive label and unmatched ones receive a negative label. At inference time, we threshold on the predicted logits to output only valid planes.

\subsection{Symmetry plane fitting}

Our model is trained on dense signed distance map predictions without extracting explicit symmetry planes. Only at inference time do we use the predicted point maps and signed distance maps to extract the actual planes.

We first filter symmetry predictions by thresholding on the predicted classification logits. For each valid plane instance, we select a high-confidence set of 3D points by thresholding on predicted confidence maps. This filtering produces a final set of predicted 3D points \(\predpoints'\) and their corresponding signed distances \(\predsdfmap'\) for each plane. We fit an optimal plane \(\predplane = (\prednormal, \predoffset)\) by solving a constrained least-squares optimization problem:
\begin{equation}
    (\prednormal, \predoffset) = \argmin_{|\normal|=1, \offset} \sum_{\point \in \predpoints', \sdf \in \predsdfmap'} \left( (\normal^\T \point + \offset) - \sdf \right)^2,
\end{equation}
which yields the final set of predicted planes.

%% file: sections/results.tex
\label{sec:detection}

We compare our method to a recent state-of-the-art single-view symmetry detector, \ssb \citep{li2025symmetry}, as well as a simple baseline, \direct, where we use features from the frozen VGGT backbone to directly regress symmetry plane parameters. 
We empirically find that parameterizing each symmetry plane by its normal direction and a point on the plane (instead of its offset) leads to more stable training and more accurate predictions.
Since \ssb was originally trained on object-level datasets without complex backgrounds (e.g., ShapeNet and Objaverse), we finetune their model on the \archsym dataset to ensure a fair comparison.

\medskip\noindent\textbf{Experimental setup.} We randomly split the 93 scenes from the \archsym dataset into a set of 74 training scenes and a set of 19 test scenes. We split by scene rather than by image to ensure that there is no scene overlap between the training and test sets. This is crucial for preventing data leakage
and for accurately evaluating generalizability to unseen structures. 

\medskip\noindent\textbf{Evaluation metrics.} Since \ssb predicts symmetry plane normals while \ours and \direct predict full plane parameters, we use two different metrics for comparison.

\medskip\noindent\textit{Normal-only.} Following \citet{li2025symmetry}, we use two evaluation metrics based on the geodesic distance, or angular error, of the predicted normals. 
For a given input image \(\image\), we denote the set of ground truth symmetry plane normals as \(\gtnormals\), the set of predicted plane normals as \(\prednormals\), and the set of ground truth symmetries that are \emph{visible} from the input image
as \(\gtnormalsvis \subseteq \gtnormals\). Specifically, a symmetry plane is considered visible in an image if at least 5\% of all visible points lie on each side of the plane. This filtering is necessary as a zoomed-in view of one facade, for example, provides no visual evidence for symmetries on other distinct facades of the building. We refer to the supplementary material for details on visibility filtering.

The geodesic distance is computed as the average of two metrics: 
\emph{exactness}, defined as the mean angular error from each predicted \(\prednormal \in \prednormals\) to the closest ground truth \(\normal \in \gtnormals\), 
and \emph{completeness}, defined as the mean angular error from each \emph{visible} ground truth \(\normal \in \gtnormalsvis\) to the closest predicted \(\prednormal \in \prednormals\). 
Exactness is assessed against all valid symmetries while completeness is assessed only against symmetries visible in the image. 
We also report the F-score at various angular thresholds (F@\(x^\circ\)), which reflects the proportion of correct predictions within the threshold. We refer to the supplementary material for details on F-score calculation.

\begin{table}[t]
\centering
\caption{\textbf{Quantitative evaluation of single-view symmetry detection.} Evaluation metrics are averaged across all test scenes. \ssb~\citep{li2025symmetry} and \direct are abbreviated as \ssbshort and \directshort. Geo: geodesic distance ($\downarrow$, degrees). F@$x^\circ$: F-score at $x^\circ$ threshold ($\uparrow$). $\mathcal{E}_{\text{dense}}$: dense symmetry error ($\downarrow$).}
\vspace{-0.5em}
\label{tab:detection_quantitative}
\small %
\setlength{\tabcolsep}{4pt} %
\begin{tabular}{@{}lccccc@{}} %
\toprule
\multirow{2}{*}{Method} & \multicolumn{4}{c}{Normal-only} & Full-plane \\
\cmidrule(lr){2-5} \cmidrule(lr){6-6}
& Geo $\downarrow$ & F@1$^\circ \uparrow$ & F@5$^\circ \uparrow$ & F@15$^\circ \uparrow$ & $\mathcal{E}_{\text{dense}} \downarrow$ \\
\midrule
\ssbshort & 10.46 & 0.07 & 0.34 & 0.55 & --- \\
\directshort & 5.06 & 0.16 & 0.64 & 0.81 & 0.18 \\
\ours & \textbf{3.71} & \textbf{0.25} & \textbf{0.70} & \textbf{0.84} & \textbf{0.13} \\
\bottomrule
\end{tabular}
\vspace{-1em}
\end{table}
\medskip\noindent\textit{Full-plane.} We follow \citet{shi2020symmetrynet} and report the dense symmetry error defined with respect to the ground truth point cloud \(\points\). 
We denote the reflection transformation of a plane \(\plane\) as \(\reflection_\plane\). Then, the dense symmetry error is computed as
\begin{equation}
    \error(\predplane, \plane) = \frac{1}{|\points|}\sum_{\point\in\points}\frac{\|\reflection_\predplane(\point) - \reflection_\plane(\point)\|}{\rho(\points, \plane)},
\end{equation}
where \(\rho(\points, \plane)\) is a normalization constant equal to the maximum distance from any point in \(\points\) to the ground truth plane \(\plane\).
Using the pairwise error as our distance measure, the final error is computed by averaging the exactness and completeness components, analogous to the geodesic distance metric in the normal-only evaluation.

\medskip\noindent\textbf{Results.} We summarize our single-view symmetry detection evaluation results, averaged across 19 test scenes, in Table \ref{tab:detection_quantitative}. As shown in the table, \ours outperforms both \ssb and \direct in normal-only prediction and outperforms \direct in full-plane prediction. We refer to the supplementary material for detailed per-scene evaluation results.

\medskip\noindent\textbf{Qualitative comparisons.} We show qualitative comparisons on eight test scene images in Figure~\ref{fig:detection_qualitative}. While \ssb can often identify the orientation of the most dominant symmetry plane, it struggles with partially visible symmetries and produces redundant detections. Meanwhile, \direct benefits from the VGGT backbone features and predicts more accurate normals, but often predicts planes misaligned with the scene geometry, highlighting the challenge of directly regressing plane parameters. In contrast, our signed-distance-based parameterization allows \ours to consistently predict planes that are well-aligned with the scene geometry. 

\medskip\noindent\textbf{Applications.} 
We demonstrate the accuracy of our symmetry detector via a simple downstream application in single-view point cloud completion in Figure~\ref{fig:placeholder1}. 
Since geometric foundation models like VGGT~\citep{wang2025vggt} only predict geometry for visible pixels in the image, output point clouds are inherently incomplete (e.g., missing the back of a building). 
By simply reflecting the predicted points across our detected symmetry planes, we can accurately ``hallucinate'' the occluded geometry.

%% file: sections/limitations.tex
Our approach to curating \archsym relies on the availability of accurate SfM reconstructions, whose quality directly influences the extracted symmetries.
While our current pipeline focuses exclusively on reflectional symmetries, it can be adapted to extract rotational symmetries by matching images against \emph{non-reflected} versions of other views. We did not prioritize this as pure rotational symmetries are less prevalent in architectural landmarks and can often be derived from the composition of reflectional symmetries.
Our pipeline focuses on discovering global symmetries by aggregating correspondences from full-image matching. 
It is not designed to detect partial symmetries (e.g., individual windows), which is an interesting direction for future work.

The performance of our symmetry detector is linked to the geometry predicted by the VGGT point head.
Our signed-distance-based approach prioritizes geometric consistency and may produce less accurate planes when the predicted geometry is highly ambiguous, e.g., due to severe occlusion or blurring.
While methods that directly regress plane parameters may produce plausible orientation estimates in such cases, it is unclear whether they can be meaningfully localized in 3D.

%% file: sections/conclusion.tex
\begin{figure}[t]
    \centering
    \includegraphics[width=\linewidth]{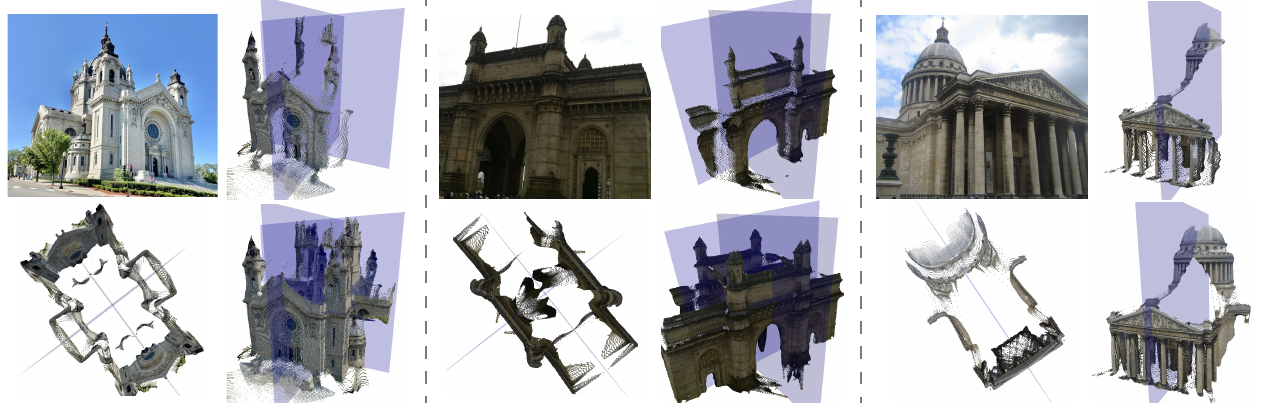}
    \vspace{-1.5em}
    \caption{Single-view completion using detected symmetries. We visualize the input image (top left), a top-down view of the completed point cloud (bottom left), and frontal view comparisons of the original VGGT (top right) and our completed results (bottom right).}
    \label{fig:placeholder1}
    \vspace{-0.5em}
\end{figure}

We present a complete pipeline for tackling the previously unaddressed problem of detecting 3D-grounded symmetries from single RGB images of in-the-wild scenes. Due to the lack of existing training data, we first introduce a novel automated pipeline that leverages cross-image reflection matching within SfM reconstructions to curate \archsym, a large-scale dataset of architectural images labeled with symmetries. Building on the dataset, we propose a novel detection model that parameterizes symmetry planes as signed distance maps relative to the model's own predicted geometry, naturally resolving the scale ambiguity issue inherent in monocular reconstruction. Our experiments demonstrate that our approach significantly outperforms existing alternatives. We believe our \archsym dataset and method provide a strong foundation for future research,
allowing 3D symmetries to be used as a robust geometric prior for in-the-wild 3D reconstruction and pose estimation.

%% file: sections/suppl.tex
\maketitlesupplementary

\begin{table*}[t]
    \centering
    \small
    \setlength{\tabcolsep}{6pt} 
    
    \begin{tabular}{l *{5}{r@{\hspace{5pt}}r@{\hspace{5pt}}r}}
    \toprule
    & \multicolumn{3}{c}{Geo $\downarrow$} & \multicolumn{3}{c}{F@1\(^\circ\) $\uparrow$} &
      \multicolumn{3}{c}{F@5\(^\circ\) $\uparrow$} & \multicolumn{3}{c}{F@15\(^\circ\) $\uparrow$} &
      \multicolumn{2}{c}{\(\mathcal{E}_{\text{dense}}\) $\downarrow$} \\
    \cmidrule(lr){2-4}\cmidrule(lr){5-7}\cmidrule(lr){8-10}\cmidrule(lr){11-13}\cmidrule(lr){14-15}
    Scene & \ssbshort & \directshort & \ours & \ssbshort & \directshort & \ours & \ssbshort & \directshort & \ours & \ssbshort & \directshort & \ours & \directshort & \ours \\ %
    \midrule
    Arc de Triomphe & 8.19 & 4.29 & \textbf{1.44} & 0.06 & 0.12 & \textbf{0.36} & 0.34 & 0.64 & \textbf{0.79} & 0.62 & 0.89 & \textbf{0.90} & 0.24 & \textbf{0.12} \\
    Arch of Hadrian & 23.17 & 3.22 & \textbf{2.00} & 0.02 & 0.13 & \textbf{0.33} & 0.34 & 0.73 & \textbf{0.78} & 0.54 & \textbf{0.92} & 0.90 & 0.19 & \textbf{0.11} \\
    Basilica of Bom Jesus & 6.93 & 1.75 & \textbf{1.48} & 0.06 & 0.22 & \textbf{0.31} & 0.30 & \textbf{0.79} & 0.77 & 0.56 & 0.84 & \textbf{0.86} & 0.07 & \textbf{0.04} \\
    Bath Abbey & 5.30 & 3.10 & \textbf{1.93} & 0.08 & 0.18 & \textbf{0.32} & 0.35 & 0.61 & \textbf{0.64} & 0.51 & \textbf{0.81} & 0.79 & 0.11 & \textbf{0.08} \\
    Cathedral of Saint Paul & 8.84 & \textbf{7.54} & 7.78 & 0.01 & 0.08 & \textbf{0.08} & 0.25 & \textbf{0.47} & 0.37 & 0.50 & 0.81 & \textbf{0.85} & 0.25 & \textbf{0.19} \\
    Charlottenburg Palace & 15.43 & \textbf{1.48} & 1.86 & 0.12 & \textbf{0.29} & 0.17 & 0.25 & \textbf{0.88} & 0.83 & 0.34 & \textbf{0.94} & 0.85 & \textbf{0.05} & \textbf{0.05} \\
    Frauenkirche (Dresden) & 13.05 & 12.79 & \textbf{9.18} & 0.01 & 0.02 & \textbf{0.06} & 0.15 & 0.28 & \textbf{0.55} & 0.45 & 0.64 & \textbf{0.74} & 0.38 & \textbf{0.27} \\
    Gateway of India & 8.35 & 5.70 & \textbf{4.89} & 0.06 & 0.11 & \textbf{0.21} & 0.35 & 0.60 & \textbf{0.61} & 0.56 & \textbf{0.84} & 0.79 & 0.20 & \textbf{0.17} \\
    Illinois State Capitol & 7.67 & \textbf{1.82} & 2.06 & 0.13 & \textbf{0.30} & 0.27 & 0.43 & 0.77 & \textbf{0.84} & 0.58 & 0.84 & \textbf{0.92} & 0.10 & \textbf{0.08} \\
    Isa Khan Niyazi's tomb & 10.48 & \textbf{8.24} & 8.44 & 0.03 & 0.04 & \textbf{0.05} & 0.23 & \textbf{0.38} & 0.36 & \textbf{0.68} & 0.64 & 0.62 & 0.29 & \textbf{0.25} \\
    Montmartre & 3.66 & 1.77 & \textbf{0.92} & 0.16 & 0.28 & \textbf{0.52} & 0.62 & 0.80 & \textbf{0.86} & 0.76 & 0.88 & \textbf{0.91} & 0.07 & \textbf{0.04} \\
    Notre-Dame Basilica & 7.79 & 2.27 & \textbf{1.79} & 0.10 & 0.20 & \textbf{0.31} & 0.40 & \textbf{0.75} & 0.73 & 0.61 & 0.88 & \textbf{0.90} & 0.08 & \textbf{0.05} \\
    Panthéon de Paris & 7.88 & 1.89 & \textbf{1.34} & 0.21 & 0.29 & \textbf{0.45} & 0.50 & \textbf{0.88} & 0.85 & 0.63 & \textbf{0.93} & 0.91 & 0.09 & \textbf{0.05} \\
    Royal Liver Building & 8.32 & 4.99 & \textbf{2.59} & 0.05 & 0.05 & \textbf{0.21} & 0.34 & 0.54 & \textbf{0.73} & 0.60 & 0.85 & \textbf{0.87} & 0.22 & \textbf{0.11} \\
    Saints Peter and Paul Church & 6.38 & \textbf{1.85} & 1.92 & 0.05 & 0.27 & \textbf{0.28} & 0.46 & \textbf{0.88} & 0.87 & 0.66 & 0.92 & \textbf{0.96} & 0.07 & \textbf{0.06} \\
    Torre de Belém & 14.55 & 12.54 & \textbf{7.17} & 0.01 & 0.04 & \textbf{0.09} & 0.16 & 0.33 & \textbf{0.64} & 0.39 & 0.65 & \textbf{0.79} & 0.35 & \textbf{0.20} \\
    Town Hall Tower in Kraków & 16.14 & 14.83 & \textbf{9.73} & 0.01 & 0.03 & \textbf{0.12} & 0.13 & 0.24 & \textbf{0.60} & 0.34 & 0.54 & \textbf{0.74} & 0.44 & \textbf{0.28} \\
    Victoria Memorial & 7.86 & \textbf{2.20} & 2.50 & 0.07 & 0.21 & \textbf{0.24} & 0.40 & \textbf{0.80} & 0.63 & 0.55 & \textbf{0.89} & 0.84 & \textbf{0.08} & 0.10 \\
    Westminster Abbey & 8.70 & 1.91 & \textbf{1.38} & 0.07 & 0.23 & \textbf{0.37} & 0.38 & 0.78 & \textbf{0.80} & 0.56 & 0.87 & \textbf{0.88} & 0.09 & \textbf{0.07} \\
    \midrule
    Mean & 10.46 & 5.06 & \textbf{3.71} & 0.07 & 0.16 & \textbf{0.25} & 0.34 & 0.64 & \textbf{0.70} & 0.55 & 0.81 & \textbf{0.84} & 0.18 & \textbf{0.13} \\
    \bottomrule
    \end{tabular}
    \caption{\textbf{Detailed per-scene comparison across three methods.} For each scene, we report the median geodesic distance and dense symmetry error, which are robust against outlier images. \ssb~\citep{li2025symmetry} and \direct are abbreviated as \ssbshort and \directshort. Geo: geodesic distance ($\downarrow$, degrees). F@$x^\circ$: F-score at $x^\circ$ threshold ($\uparrow$). $\mathcal{E}_{\text{dense}}$: dense symmetry error ($\downarrow$).}
    \label{tab:detection_quantitative_full}
\end{table*}

\section{Implementation details}

Our implementation builds upon the official MASt3R~\cite{duisterhof2024mast3r} and VGGT~\citep{wang2025vggt} codebases.

\subsection{Training details}
We use a base learning rate of 1e-4 and a cosine decay learning rate schedule, with an effective batch size of 48. For data augmentation, we perform random center crop, random horizontal flip, and color jitter. Training is performed on 4 A6000 GPUs for 2 days.

\subsection{Network architecture}
Our model consists of the pre-trained VGGT~\citep{wang2025vggt} model and our symmetry prediction head. The VGGT backbone and point prediction head are frozen with weights from the officially released checkpoints. We find no significant difference in performance between using the point head and using the depth and camera heads for point map prediction. 

The architecture of our symmetry prediction head is similar to the VGGT point prediction head, with the only difference being the additional FiLM conditioning~\citep{perez2018film} before each fusion block. We use eight instance queries to identify up to eight reflectional symmetries. They are passed through a three-layer transformer decoder with an embedding dimension of 256. The instance queries attend to the final layer features of the VGGT backbone, which are projected to the same embedding dimension. The refined instance queries are passed through a two-layer MLP to obtain four pairs of FiLM conditioning parameters for each of the four fusion blocks.

Classification logits are predicted by a two-layer MLP that takes in the final upsampled feature maps after global average pooling. We find no significant difference in performance between regressing classification logits directly from the refined instance queries and from the final upsampled feature maps.

\begin{algorithm}[b]
\caption{Visibility filtering heuristic}
\label{alg:filtering}
\begin{algorithmic}[1]
\Require Depth map \(\depthmap\), camera parameters \((\intrinsics, \extrinsics)\), plane parameters \(\plane = (\normal, \offset)\)
\Ensure Boolean indicating if the plane \(\plane\) is visible.

\State Crop \(\depthmap\) to its central 80\% region, yielding \(\depthmap_c\).
\State Let \(\mathcal{X}_c\) be the set of pixel coordinates with valid depth.
\State \(N_{\text{valid}} \gets |\mathcal{X}_c|\)
\If{\(N_{\text{valid}} < 1000\)} \Comment{not enough valid pixels}
    \State \Return \textbf{False}
\EndIf

\State Initialize \(N_{\text{pos}} \gets 0\), \(N_{\text{neg}} \gets 0\)
\ForAll{pixel \(\pixel_k \in \mathcal{X}_c\)}
    \State \(\point_k \gets \mathsf{unproject}(\depthmap_c(\pixel_k), \intrinsics, \extrinsics)\).
    \State \(s_k \gets \normal^\T \point_k + \offset\).
    \If{\(s_k > 0\)} \Comment{positive signed distance}
        \State \(N_{\text{pos}} \gets N_{\text{pos}} + 1\)
    \ElsIf{\(s_k < 0\)} \Comment{negative signed distance}
        \State \(N_{\text{neg}} \gets N_{\text{neg}} + 1\)
    \EndIf
\EndFor

\State \(\text{prop}_{\text{pos}} \gets N_{\text{pos}} / N_{\text{valid}}\) \Comment{filter by proportion}
\State \(\text{prop}_{\text{neg}} \gets N_{\text{neg}} / N_{\text{valid}}\)

\If{\(\text{prop}_{\text{pos}} < 0.05\) \textbf{or} \(\text{prop}_{\text{neg}} < 0.05\)}
    \State \Return \textbf{False}
\Else
    \State \Return \textbf{True}
\EndIf

\end{algorithmic}
\end{algorithm}

\begin{algorithm}[b]
\caption{Visibility-aware F-score calculation}
\label{alg:fscore}
\begin{algorithmic}[1]
\Require Predicted normals \(\prednormals\), full GT normals \(\gtnormals\), visible GT normals \(\gtnormalsvis\), threshold \(x^\circ\)
\Ensure F-score at threshold \(x^\circ\)

\State Find optimal matching \(\matching\) between \(\prednormals\) and \(\gtnormals\) based on geodesic distance.
\State Initialize \(\tp \gets 0\), \(\fp \gets 0\), \(\fn \gets 0\), \(\ignore \gets 0\)

\ForAll{pair \((\prednormal_i, \normal_j)\) with distance \(d_{ij}\) in \(\matching\)}
    \If{\(d_{ij} < x^\circ\)} \Comment{distance within threshold}
        \If{\(\normal_j \in \gtnormalsvis\)} \Comment{matched to visible GT}
            \State \(\tp \gets \tp + 1\) 
        \Else \Comment{matched to non-visible GT}
            \State \(\ignore \gets \ignore + 1\) 
        \EndIf
    \EndIf
\EndFor 

\State \(\fp \gets |\prednormals| - \tp - \ignore\)
\State \(\fn \gets |\gtnormalsvis| - \tp\)

\State \Return \((2 \cdot \tp)/(2 \cdot \tp + \fp + \fn)\)

\end{algorithmic}
\end{algorithm}

\begin{table}[b]
\centering
\caption{\textbf{Ablation studies on model architecture and loss supervision.} Plane prediction quality severely degrades if we remove FiLM conditioning~\citep{perez2018film} (w/o FiLM) or use ground truth signed distance map supervision (w/ GT) that is inconsistent with the predicted geometry.}
\label{tab:ablation_study}
\small
\setlength{\tabcolsep}{4pt}
\begin{tabular}{@{}lccccc@{}}
\toprule
\multirow{2}{*}{Method} & \multicolumn{4}{c}{Normal-only} & Full-plane \\
\cmidrule(lr){2-5} \cmidrule(lr){6-6}
& Geo $\downarrow$ & F@1$^\circ \uparrow$ & F@5$^\circ \uparrow$ & F@15$^\circ \uparrow$ & $\mathcal{E}_{\text{dense}} \downarrow$ \\
\midrule
Full & \textbf{3.71} & \textbf{0.25} & \textbf{0.70} & \textbf{0.84} & \textbf{0.13} \\
w/o FiLM & 6.72 & 0.16 & 0.51 & 0.73 & 0.20 \\
w/ GT & 6.99 & 0.13 & 0.48 & 0.73 & 0.19 \\
\bottomrule
\end{tabular}
\end{table}

\section{Evaluation details}

Detailed per-scene evaluation results on 19 test scenes are reported in Table~\ref{tab:detection_quantitative_full}. For each scene, we report the \emph{median} geodesic distance and dense symmetry error, which are robust against outlier images within a scene (e.g., images with extreme viewpoints or severe occlusions). The mean statistics are then computed as the average across all per-scene statistics.

\subsection{Visibility filtering heuristic}

Although our dataset curation pipeline identifies ground truth symmetries at the scene level, for training and evaluation purposes, it is necessary to determine which of these symmetries are actually visible in each image. A symmetry plane is considered visible if the image captures sufficient geometric structure (e.g. 5\% of pixels with valid depth) on both sides of the plane. Additionally, we filter out images where the landmark structure is not the primary subject to ensure a clean training signal. The details of our filtering heuristic are presented in Algorithm~\ref{alg:filtering}.

\subsection{Asymmetric F-score calculation}

We present our F-score calculation scheme in Algorithm~\ref{alg:fscore}, which is based on the evaluation code from \ssb~\citep{li2025symmetry}. We modify the code to take into account our asymmetric evaluation scheme---all ground truth planes are used in evaluating \emph{exactness}, but only visible ground truth planes are used in evaluating \emph{completeness}. In particular, bipartite matching is performed between predicted planes and \emph{all} ground truth planes instead of only the visible ground truth planes. Predicted planes that are within the angular error threshold of some non-visible ground truth plane are \emph{not} considered as false positives during F-score calculation, whereas in a standard F-score calculation scheme they would be. Intuitively, this means we penalize the model for not predicting visible symmetries, but we do not penalize the model for \emph{accurately} predicting non-visible symmetries when they apparant from other indirect cues.

\section{Ablation studies}

We validate our design choices by comparing the full model against two baselines. 
First, we remove the instance-specific FiLM conditioning~\citep{perez2018film} (w/o FiLM) to predict multiple symmetry planes directly from shared DPT features~\citep{ranftl2021vision}.
Second, we supervise the model using ground truth signed distance maps (w/ GT) derived from the ground truth geometry instead of the pseudo-ground truth (derived from the model's predicted point maps) that is consistent with the predicted geometry. 
As shown in Table~\ref{tab:ablation_study}, we observe that both the orientation and alignment of the predicted planes degrade severely in both cases, highlighting the effectiveness of our two-stage model architecture and the importance of self-consistent predictions.

\section{Additional results}

\begin{figure}[t]
    \centering
    \includegraphics[width=\linewidth]{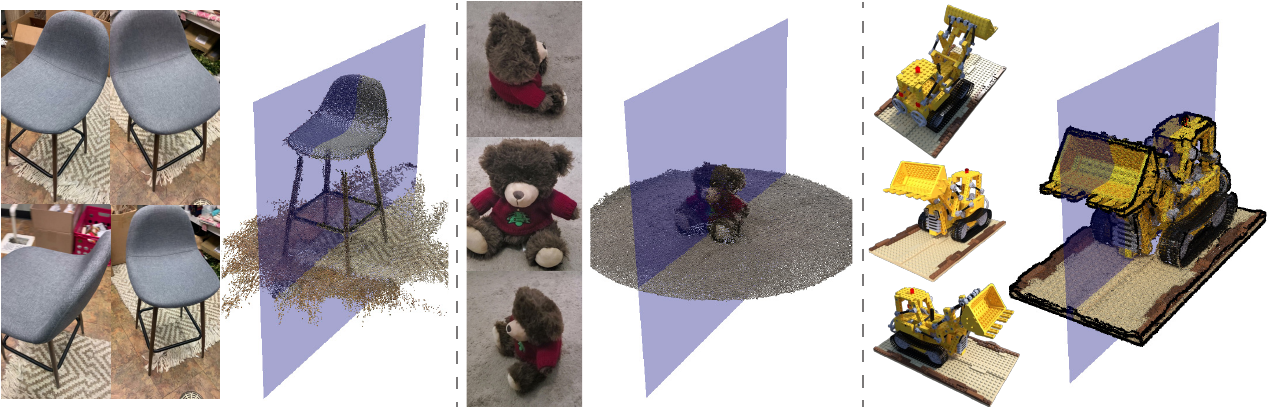}
    \caption{\textbf{Generalization to object-centric scenes.} We demonstrate accurate symmetry annotation on real (CO3D~\citep{reizenstein21co3d}, left/center) and synthetic (NeRF-Synthetic~\citep{mildenhall2020nerf}, right) objects. We show sample input images and annotated symmetry planes overlaid on COLMAP MVS~\citep{schoenberger2016mvs} point clouds.}
    \label{fig:synthetic}
    \vspace{-0.5em}
\end{figure}

\subsection{Generalization to object-centric data}

Our paper focuses on architectural scenes since the MegaScenes dataset~\cite{tung2024megascenes} already provides real-world variability (e.g., illumination, viewing angles) necessary to benchmark this task. However, our symmetry annotation pipeline and signed-distance formulation are general-purpose. As shown in Figure~\ref{fig:synthetic}, our automated annotation pipeline can correctly produce 3D symmetries when directly applied to non-architectural object-centric scenes. We run our annotation pipeline on three scenes sampled from the CO3D~\citep{reizenstein2021common} and NeRF-Synthetic~\citep{mildenhall2020nerf} datasets. The detected reflectional symmetry planes are overlaid on dense point clouds from COLMAP MVS~\citep{schoenberger2016mvs} for visualization.

\subsection{Additional qualitative comparisons}

\begin{figure*}
    \centering
    \includegraphics[width=\linewidth]{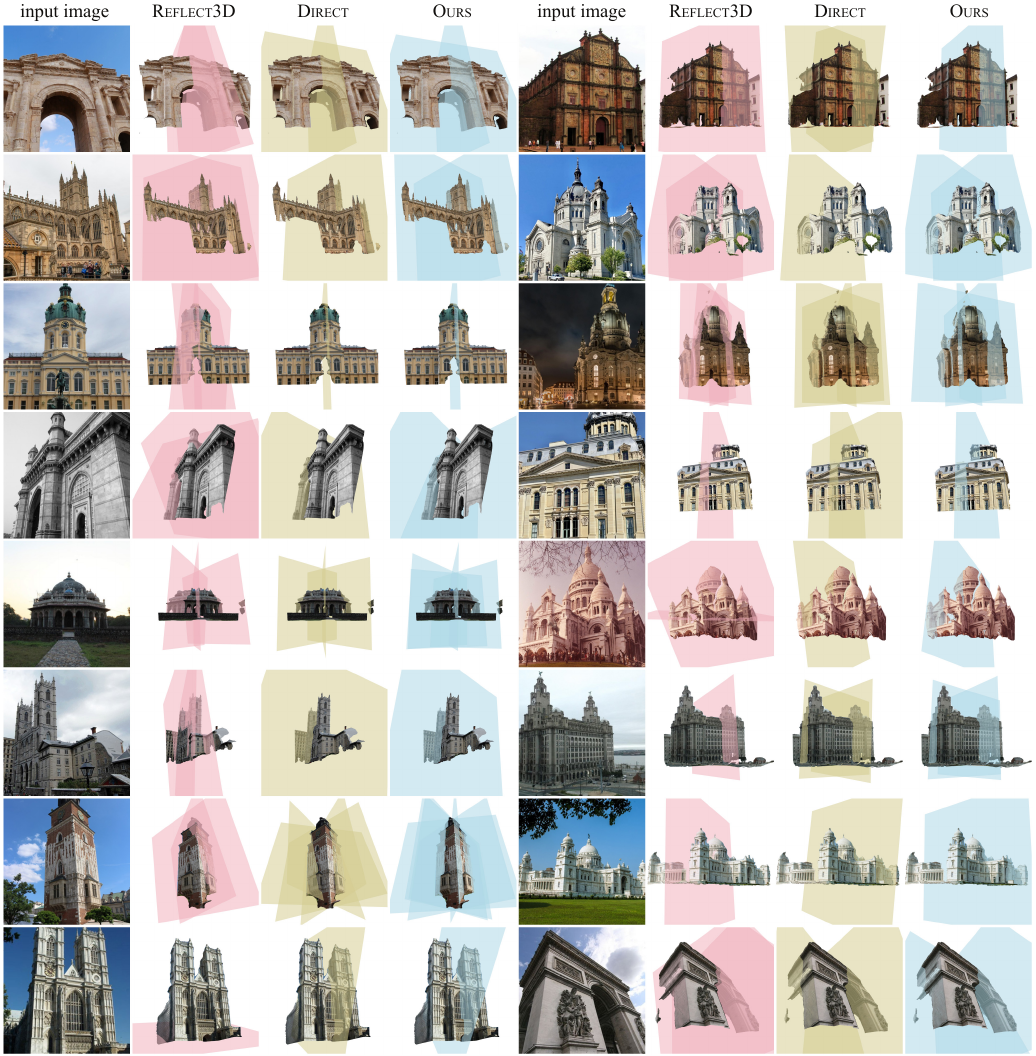}
    \caption{\textbf{Additional qualitative comparisons of single-view symmetry detection results.} Input images are sampled from 16 different test scenes. \ssb~\citep{li2025symmetry} often misses partially visible symmetries and produces redundant detections, while \direct often predicts planes that are misaligned with the scene geometry. We encourage zooming into the figure to see differences in plane orientation and alignment in detail.}
    \label{fig:additional}
\end{figure*}

Figure~\ref{fig:additional} presents additional qualitative comparisons on images sampled from 16 test scenes. These examples further illustrate that our signed-distance parameterization allows \ours to consistently predict planes that are better aligned with the underlying scene geometry.

\subsection{Per-query predictions}

\begin{figure*}
    \centering
    \includegraphics[width=\linewidth]{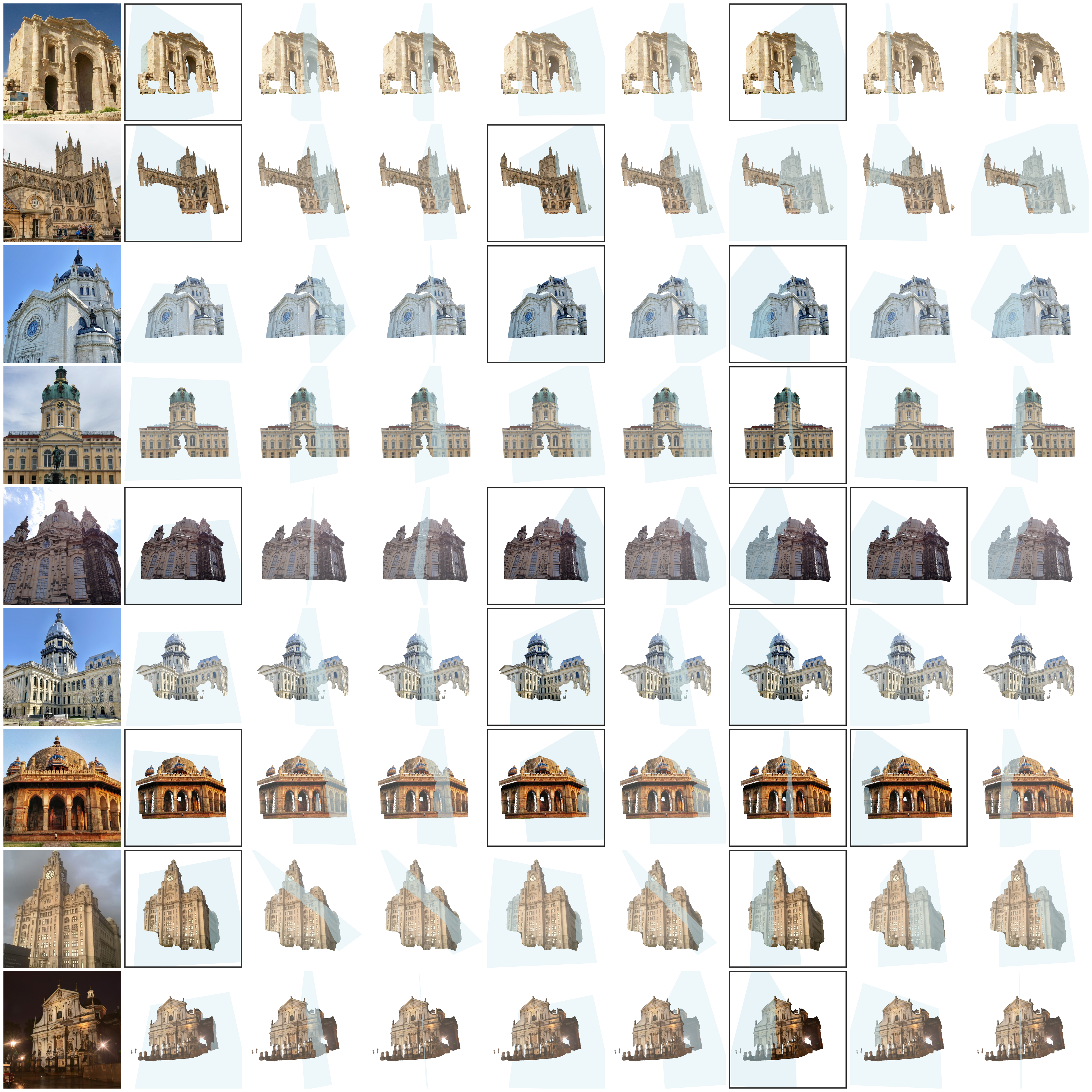}
    \caption{\textbf{Visualization of symmetry plane predictions from individual instance queries.} Each row shows an input image alongside the symmetry planes predicted from each of the eight instance queries. Highlighted frames indicate valid planes with predicted logits above the extraction threshold. This visualization demonstrates how different prediction slots specialize to capture specific types of symmetries within the scene.}
    \label{fig:instance}
\end{figure*}

To provide insight into the behavior of our multi-instance detection head, we visualize the symmetry plane predictions associated with each of the eight learnable instance queries in Figure~\ref{fig:instance}. Subplots with highlighted frames correspond to valid symmetry planes (i.e., those with logits above the extraction threshold). We observe that prediction slots corresponding to different instance queries learn to specialize in extracting different types of symmetries. For example, the first slot often detects a front-to-back reflection, while the sixth slot often detects a reflection across the main facade. Notably, this specialization can be observed even when the corresponding symmetry is not present in the specific scene (resulting in suppressed predictions). This highlights the effectiveness of our two-stage architecture  and set prediction formulation in handling scenes with varying numbers and types of symmetry planes.

